\documentclass[sigconf,screen]{acmart}

\AtBeginDocument{%
  \providecommand\BibTeX{{%
    \normalfont B\kern-0.5em{\scshape i\kern-0.25em b}\kern-0.8em\TeX}}}

\setcopyright{acmlicensed}
\copyrightyear{2025}
\acmYear{2025}
\acmDOI{XXXXXXX.XXXXXXX}

\acmConference[manuscript]{Make sure to enter the correct conference title from your rights confirmation emai}{2025}
\usepackage{amsmath}
\usepackage{booktabs}
\usepackage{makecell}
\usepackage{multirow}
\usepackage{amssymb}
\usepackage{pifont}
\newcommand{\cmark}{\ding{51}}%
\newcommand{\xmark}{\ding{55}}
\PassOptionsToPackage{table,dvipsnames}{xcolor}
\definecolor{ForestGreen}{rgb}{0.13, 0.55, 0.13}
\definecolor{AoGreen}{rgb}{0.0, 0.5, 0.0}
\definecolor{cadmiumred}{rgb}{0.89, 0.0, 0.13}
\begin{document}
\title{GraphT5: Unified Molecular Graph-Language Modeling via Multi-Modal Cross-Token Attention}

\author{Sangyeup Kim}
\authornote{Both authors contributed equally to this research.}
\email{sang2668@snu.ac.kr}
\affiliation{%
  \institution{Computer Science and Engineering, \\Seoul National University}
  \country{South Korea}
  }
  
\author{Nayeon Kim}
\authornotemark[1]
\email{ny_1031@snu.ac.kr}
\affiliation{%
  \institution{Interdisciplinary Program in Artificial Intelligence, \\Seoul National University}
  \country{South Korea}
  }

\author{Yinhua Piao}
\email{2018-27910@snu.ac.kr}
\affiliation{%
  \institution{Computer Science and Engineering, \\Seoul National University}
  \country{South Korea}
}
\author{Sun Kim}
\affiliation{%
  \institution{Interdisciplinary Program in Artificial Intelligence; Computer Science and Engineering, Seoul National University; AIGENDRUG Co., Ltd.}
  \country{South Korea}
}

\renewcommand{\shortauthors}{Kim et al.}


\begin{abstract} % 160~230 words

% 239 words 
Molecular language modeling tasks such as molecule captioning have been recognized for their potential to further understand molecular properties that can aid drug discovery or material synthesis based on chemical reactions. Unlike the common use of molecule graphs in predicting molecular properties, most methods in molecular language modeling rely heavily on SMILES sequences. This preference is because the task involves generating a sequence of multiple tokens using transformer-based models. Therefore, a main challenge is determining how to integrate graph data, which contains structural and spatial information about molecules, with text data. In addition, simply using both 1D SMILES text and 2D graph as inputs without addressing how they align and represent the molecule structure in different modalities makes it challenging to fully utilize structural knowledge about molecules. To this end, we propose GraphT5, a multi-modal framework that integrates 1D SMILES text and 2D graph representations of molecules for molecular language modeling. 
Specifically, we introduce a novel cross-token attention module in GraphT5 to bridge the gap arising from the fundamental differences between the two modalities of molecule representations. 
Cross-token attention exploits implicit information between SMILES and graphs of molecules, resulting from their interactions at a fine-grained token level that benefits molecular language modeling.
Extensive experiments including molecule captioning, IUPAC name prediction tasks, and case studies show that our GraphT5 outperforms the latest baseline approaches, which validates the effectiveness of our GraphT5 in sufficiently utilizing 1D SMILES text and 2D graph representations.
\end{abstract}





\maketitle

\section{Introduction}
\begin{figure}[t]
\centering
\includegraphics[width=1\columnwidth]{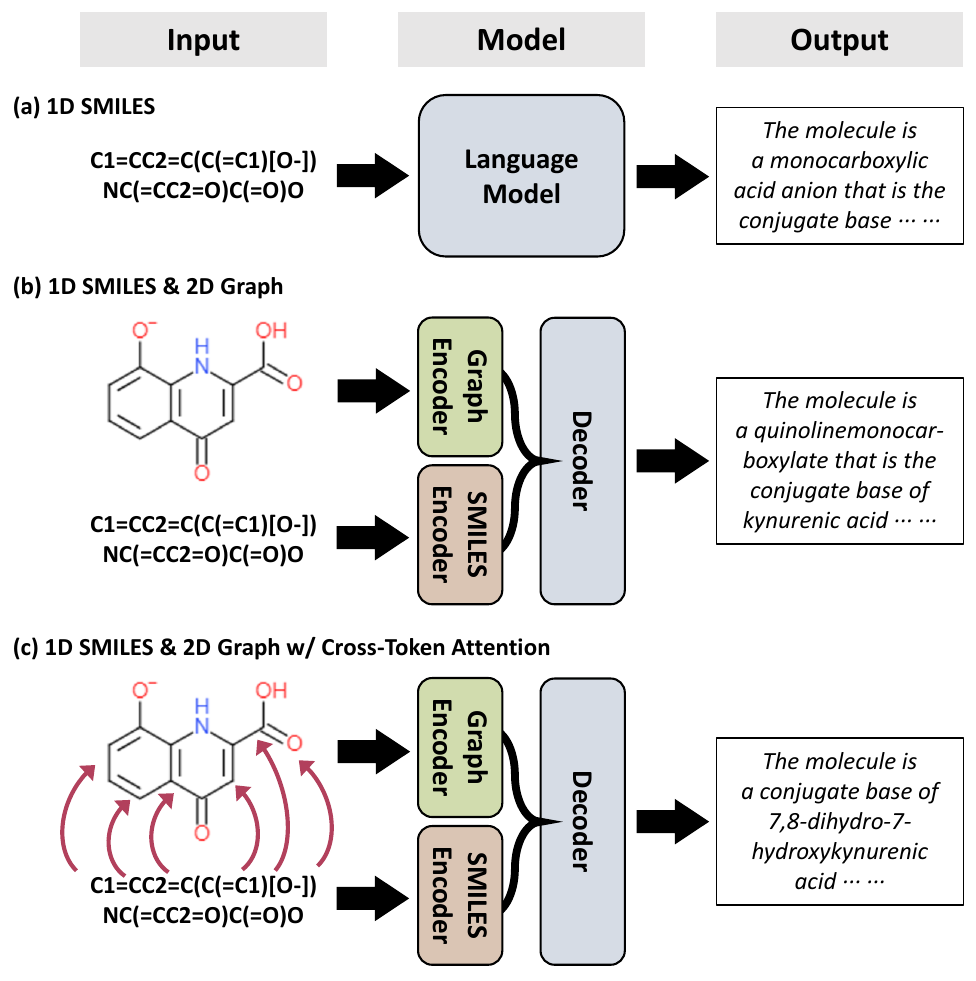} 
\caption{Molecule captioning with different input modalities and encoders. (a) 1D SMILES (Simplified Molecular Input Line Entry System) \cite{weininger1988smiles} with language model (e.g. T5-based model). (b) 1D SMILES and 2D graph as input for graph and SMILES encoders, using text decoder. (c) 1D SMILES and 2D graph with cross-attention between SMILES and graph as input for graph and SMILES encoders, using text decoder.}
\label{fig:summary}
\end{figure}

As deep learning technology advances, various models have emerged in the field of chemistry \cite{baum2021artificial}, particularly in molecular domain. The tasks that molecular deep learning models address include predicting various properties of molecules, such as Blood-Brain Barrier Penetration (BBBP), lipophilicity, and others\cite{fang2022geometry}. These developments are significant as they hold immense potential in aiding key areas of chemistry like drug discovery and material synthesis. As the field of molecular deep learning continues to expand, diverse methods for processing and interpreting molecular data have emerged. Molecules can primarily be depicted in two distinct forms: as textual representations, often using sequences like SMILES \cite{weininger1988smiles}, and as graph representations \cite{david2020molecular} that molecules are depicted as graphs where atoms are nodes and bonds are edges. Initially, most approaches relied on SMILES for molecular deep learning, but over time, research has increasingly shown that graph-based approaches are more effective. Graph-based approaches capture the spatial relationships and structural features of molecules. Deep learning models, especially Graph Neural Networks (GNNs) \cite{scarselli2008graph} based models, have shown exceptional performance in the fields including predicting molecular properties and 3D conformers, surpassing models that utilize SMILES data. In other words, predicting molecule properties such as label predictions or regressions have been successfully modeled with 1D SMILES, 2D graphs, and also 3D representations of molecules. Recently, molecular language modeling tasks such as molecule captioning\cite{edwards2022translation} and IUPAC name prediction\cite{iupac2005nomenclature} have attracted considerable attention, primarily due to their potential applications in the fields of chemistry and biology. As the tasks often involve generating sequences of multiple tokens to describe molecules, the majority of current approaches employ language models with one-dimensional SMILES data as input, as depicted in Figure \ref{fig:summary} (a). However, relying solely on 1D text representation limits the reflection of structural and spatial information to generate multiple text tokens. The key challenge is to integrate this 1D text representation with the 2D structural information of molecules remains in this approach.

Simultaneously with advancements in the molecular domain, the field of computer vision has witnessed a surge in multi-modal models for image and video captioning \citep{ghandi2023deep, abdar2023review}. In these approaches, it was found that allowing cross-modal interactions between images and texts led to significant improvements in performance\cite{li2022mplug}. These developments also highlight how multi-modal approaches can be effectively utilized in captioning tasks, demonstrating their potential in various applications. Recent researches \cite{liu2023molca, su2022molecular} in molecular language modeling have suggested using both graph and text data (SMILES) in a multi-modal approach like Figure \ref{fig:summary}(b). However, this does not yet include the integration of cross-modal interactions as illustrated in Figure 1(c). Cross-modal interactions are significant since both SMILES and graph data possess unique and complementary information about molecules. Through cross-modal interaction, these distinct data types can enrich the overall understanding, allowing for a more comprehensive and accurate interpretation of molecular structures and properties.

Therefore, we introduce GraphT5, a model that integrates molecular graph data with SMILES text for molecular language modeling. This multi-modal approach leverages flat and structural information of both 1D SMILES  and 2D molecular graph representation to provide a more accurate molecular understanding. In GraphT5, we propose a novel method called `Cross-Token Attention' to capture token-level interactions between the two modalities. Then the learned multi-modal representations are fed into the T5 decoder to generate more insightful and domain-specific captions for molecules. We conduct extensive experiments on benchmark datasets to demonstrate that our model outperforms state-of-the-art methods. Results in molecule captioning and IUPAC name prediction validate the effectiveness of our approach in improving text generation accuracy and detail. In the Molecule Captioning task, GraphT5 surpasses the baselines in BLEU-2 and BLEU-4 scores with the ChEBI-20 dataset \cite{edwards2021text2mol} and demonstrates state-of-the-art performance across all scores on the PubChem324k dataset. For the IUPAC Name Prediction task, GraphT5 also shows state-of-the-art results on the PubChem324k dataset. Furthermore, our ablation experiments demonstrate the effectiveness of using graph data in conjunction with cross-token attention.

To summarize, the key contributions of this paper are as follows:
\begin{itemize}
    \item We introduce GraphT5, a multi-modal framework developed to integrate graph data in molecular language modeling tasks, enhancing the depth and accuracy of the generated text.
    \item We propose a novel method named `Cross-Token Attention' enabling GraphT5 to capture cross-modal interactions between graph and text data of molecules.
    \item We conduct extensive experiments on benchmark datasets for molecular language modeling tasks, such as molecule captioning and IUPAC name prediction, that our approach outperforms compared to baseline models. This validates the effectiveness of GraphT5 in improving the quality of molecular understanding.
\end{itemize}

\section{Related Works}
\subsection{Molecule Representation Learning}

Achieving more accurate representations of molecules is fundamental for the efficacy of molecular AI models, impacting their ability to predict and interpret complex molecular interactions and properties \cite{david2020molecular}. Initially, Molecules are often converted to text data using notations like SMILES. These notations provide a way to represent the structure of a molecule in a linear, text-based format. Since SMILES are text representations, they can be effectively processed using language models like CNN \cite{o2015introduction}, RNN \cite{sherstinsky2020fundamentals}, or Transformer \cite{Vaswani:2017:NeurIPS}. This compatibility offers an advantage in efficiently understanding the complex patterns in molecular data.

However, text-based molecule representations are limited in that they cannot accurately capture the structural information of a molecule. Therefore, Graph Neural Network (GNN) \cite{scarselli2008graph} has become increasingly popular for their ability to capture either the topological (2D GNN) or spatial (3D GNN) aspects of molecular structures \cite{liu2021pre}. While 2D GNN emphasizes node adjacency and topological relationships, 3D GNN extends this by incorporating spatial positioning of atoms, offering insights into molecular energy. GraphMVP \cite{liu2021pre} is a model-agnostic framework that can be applied to both 2D and 3D GNN representation functions. GraphMVP employs self-supervised learning (SSL) \cite{liu2021self, liu2022graph} by capitalizing on the correspondence and consistency between 2D topological structures and 3D geometric views. This results in a 2D molecular graph encoder that is significantly enhanced by the incorporation of 3D geometry, which provides a more discriminative and enriched representation of molecules.

\begin{figure*}[t]
\centering
\includegraphics[width=1\textwidth]{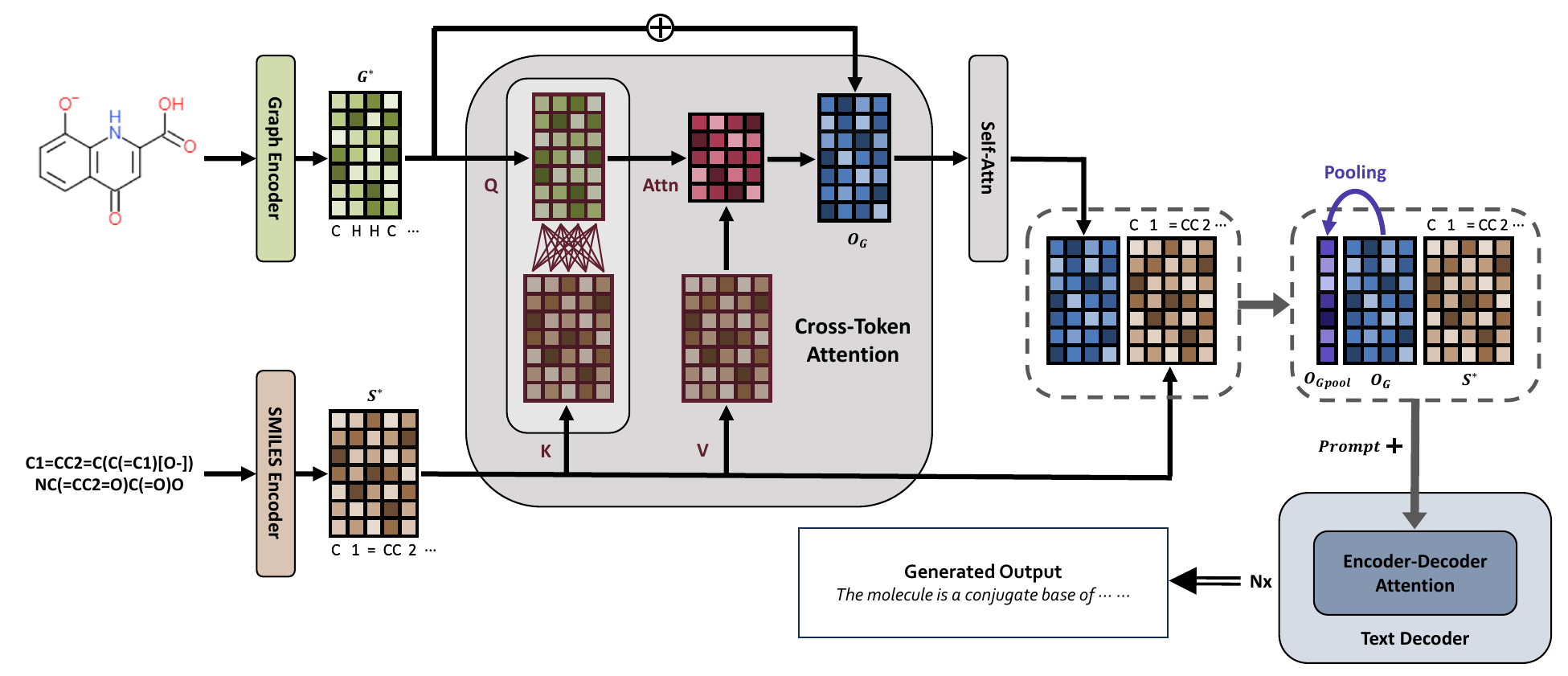} % Reduce the figure size so that it is slightly narrower than the column.
\caption{Overview of the proposed GraphT5. 1D SMILES text and 2D graph representations of the given molecule are fed into the SMILES encoder and graph encoder respectively. The following cross-token attention leverages the 1D SMILES and 2D graph representations of the molecule, resulting in token-level interaction reflected in the graph embeddings. After cross-token attention, residual connection and self-attention mechanisms are applied. The output graph embedding is summarized into a single vector by mean-pooling operation. The context vector for encoder-decoder attention in the text decoder is composed of the summarized graph vector, graph embeddings with cross-token attention applied, and the original SMILES embeddings. From the decoder, a caption of the given molecule is generated.}
\label{fig:model}
\end{figure*}

\subsection{Molecular Language Modeling}
In the field of molecular language modeling, which involves research like molecule captioning that utilizes text alongside molecular data, recent developments have shown significant progress \cite{hastings2016chebi}. This area focuses on understanding and generating textual descriptions or properties of molecules, integrating linguistic and molecular insights. These tasks can be simply addressed using language models like BERT \cite{devlin2018bert}, T5 \cite{raffel2020exploring}, or GPT \cite{radford2018improving}. Early endeavors with SMILES data include models like KV-PLM \cite{zeng2022deep}, which uses a BERT backbone and applies masked language modeling loss on 1D SMILES for molecular modeling. Another notable model is MolT5 \cite{edwards2022translation}, which is based on the T5 backbone. MolT5 represents a significant step in this direction, offering T5-based Language Models (LMs) for translations between SMILES-to-text and text-to-SMILES, and is jointly trained on molecule SMILES and a general text corpus. Building on these single-modal studies, the Text+Chem T5 model \cite{christofidellis2023unifying} marks further progress. It advances the scope beyond MolT5's focus on SMILES-text translations by facilitating cross-domain translations between chemical and natural language, thus enhancing the multi-tasking capabilities and extending the application range of the initial SMILES-based approaches.

Recently, there have been a few researches on molecular understanding that employed multi-modal approaches. MoleculeSTM \cite{liu2022multi} serves as a multi-modal foundation model that integrates molecular structural information with textual knowledge. Molecule\\-STM is designed for jointly learning molecular graphs and text, using a contrastive learning strategy. This approach to combining molecular structural data with textual descriptions underscores the potential of leveraging diverse data types for a comprehensive understanding of molecular phenomena. However, in the case of the MoMu model \cite{su2022molecular}, which attempts to encapsulate molecules via contrastive learning of graph representation and text representations, the results indicated a lower performance compared to approaches that utilize only text data. MolCA \cite{liu2023molca} utilized Q-Former to learn molecule text and graph data together, and achieved high performance in the molecule captioning task. However, they have not yet proposed a method for successfully learning cross-modal interaction. Therefore, we propose GraphT5, which jointly learns text and graph data of a molecule through cross-token attention to capture interactions between different modalities and generate an accurate molecule representation.

\section{Task Definition}
To evaluate the performance of GraphT5 in molecular language modeling, we utilize two tasks: \textit{molecule captioning} \cite{edwards2022translation} and \textit{IUPAC name prediction} \cite{iupac2005nomenclature}. Both molecular language modeling tasks use a molecule as input to generate text which is composed of multiple tokens.

For \textit{molecule captioning}, the task requires generating descriptive text for molecular structures when given a molecule as input. Examples of this can be seen in Figure \ref{fig:summary}. To assess this task more broadly, we employ two benchmark datasets: ChEBI-20 dataset \cite{edwards2021text2mol} and PubChem324k dataset. We use ChEBI-20 dataset, commonly used for benchmarking, and also conduct experiments on the PubChem324k dataset, which has longer average descriptions, adding complexity to the molecule captioning task.

\textit{IUPAC name prediction} task, as the name suggests, involves predicting the IUPAC name of a molecule. As seen in Figure \ref{fig:iupac} below, IUPAC names directly reflect the compositional and structural characteristics of molecules. This task allows for measuring how accurately the model can generate the structural features of molecules. Moreover, since generating IUPAC names is not a typical language modeling task—given the unique syntax of chemical nomenclature—it also evaluates how well the model can produce chemically specialized descriptions. Thus, the IUPAC name prediction task provides an assessment of the model's ability to generate precise and domain-specific text.

\section{Method}
\subsection{Model Architecture}
% \begin{itemize}
%     \item Detailed explanation of how prompts tokens, graph tokens, and SMILES tokens are processed together in the model to generate text.
% \end{itemize}
In this work, we propose GraphT5, a new architecture engineered for the joint learning of graph- and text-based molecular representations (Figure \ref{fig:model}). GraphT5 consists of T5 backbone \cite{raffel2020exploring} (Section \ref{method:lm}), graph encoder (Section \ref{method:graph}), and Cross-Token Attention (Section \ref{method:cross}) between molecular graph and SMILES text representations \cite{weininger1988smiles}.

\subsection{Language Model}\label{method:lm}
% \begin{itemize}
%     \item Explanation of Text+Chem T5 model.
% \end{itemize}

%%% T5 backbone 관련 추가하기 => Done! %%%
In GraphT5, we use a T5 backbone, a transformer-based encoder-decoder structure that is capable of sequence-to-sequence modeling such as machine translation or summarization in natural language processing. Moreover, since the encoder and decoder can be trained for different modalities respectively, T5 backbone can be used for cross-domain or multi-modal tasks \cite{raffel2020exploring}. Therefore, T5-based architectures are dominantly used for molecular language modeling tasks that focus on 1D SMILES representation of the molecule. In GraphT5, we employ Text+Chem T5  \citep{christofidellis2023unifying} as the SMILES text encoder and text decoder. Text+Chem T5 is a multi-domain, multi-task model using T5 backbone for forward reaction prediction, retrosynthesis, and molecular captioning tasks. Text+Chem T5 is pre-trained on ChEBI-20 \cite{edwards2021text2mol} train dataset with only molecular descriptions and fine-tuned for molecular language modeling with 1D SMILES and description pairs of ChEBI-20 train dataset. Since the text encoder of Text+Chem T5 can process SMILES representation of molecules, it is used for the SMILES text encoder of our GraphT5. Given a SMILES representation of the molecule $\mathcal{S}$, the text encoder $f$ encodes tokenized SMILES representation as:

\begin{equation}
f(\mathcal{S}) = \mathcal{S}^* \in\mathbb{R}^{|n|\times{d}}
\end{equation}
% \normalsize
where $|n|$ denotes the fixed length of the SMILES input, and $d$ denotes the dimension of the token embedding.

% Decoder
We also employ the decoder from the Text+Chem T5 model \citep{christofidellis2023unifying}. Unlike the language models for molecule captioning, the decoder of GraphT5 receives a concatenated input comprising the encoded prompt \( P \), the average pooled and original format of the output from the \textit{Cross-Token Attention} $( O_{Gpool}, O_G )$, and the encoded SMILES tokens \(\mathcal{S}^*\) along with padding. 
The concatenated input for the GraphT5 decoder is constructed as:

\begin{equation}
I_{\text{concat}} = [P, O_{Gpool}, O_G, \mathcal{S}^*]
\end{equation}
\begin{table}[t]
    \centering
    \resizebox{1\columnwidth}{!}{\begin{tabular}{cccccc}
        \toprule
        Dataset & Split & Size & Avg mol len & Min text len & Avg text len \\
        \midrule
        \multirow{3}*{PubChem324k}
        & Train & 12000 & 32 & 20 & 60 \\
        ~ & Valid & 1000 & 32 & 20 & 61 \\
        ~ & Test & 2000 & 31 & 20 & 60 \\
        \hline
        \multirow{3}*{ChEBI-20} 
        & Train & 26407 & 32 & 21 & 43 \\
        ~ & Valid & 3301 & 32 & 21 & 43 \\
        ~ & Test & 3300 & 31 & 21 & 43 \\
        \bottomrule
    \end{tabular}}
    \caption{Statistics of PubChem324k and ChEBI-20 datasets. Molecule length denotes the number of atoms that form the molecule. Text length denotes the length of the description, which is counted by splitting at spaces.}
    \label{table:dataset}
\end{table}

\begin{table}[t]
    \centering
    % \small
    %\resizebox{1\columnwidth}{!}{
    \begin{tabular}{ccc}
        \toprule
        Input & Model & \# Params\\
        \midrule
        \multirow{5}*{1D SMILES}
        & MolT5, $Small$ & 80M \\
        ~ & MolT5, $Base$ & 250M \\
        ~ & MolT5, $Large$ & 780M \\
        ~ & Text+Chem T5 & 220M \\
        ~ & Text+Chem T5, $augm$ & 220M \\
        \midrule
          & MoMu, $Base$ & 252M \\
        1D SMILES & MoMu, $Large$ & 782M \\
        $+$ & MolCA, $Galac_{125M}$ & 222M \\
        2D Graph & MolCA, $Galac_{1.3B}$ & 110M (LoRA) \\
          & GraphT5 (Ours) & 272M \\
        \bottomrule   
    \end{tabular}%}
    \caption{The number of trainable parameters of the baseline models and our GraphT5, which shows the size of the model.}
    \label{table:modelsize}
\end{table}

\subsection{Graph Encoder}\label{method:graph}
% \begin{itemize}
%     \item Explanation of using 5-layer GIN.
%     \item GraphT5 enables the implicit incorporation of molecule 3D information through using GraphMVP's pre-trained weight for graph encoder.
% \end{itemize}
In GraphT5, we employ a pre-trained Graph Isomorphism Network (GIN) \cite{xu2018powerful} to encode the graph representation of molecules. We utilize a multi-view pre-training framework, GraphMVP \cite{liu2021pre}, that leverages the correspondence and consistency between 2D topological structure and 3D geometric view. Thus, our model can indirectly encompass 3D molecular geometry. Notably, GraphMVP is pre-trained on the GEOM dataset \cite{axelrod2022geom} which contains 250,000 molecular conformations. 

A molecular graph $\mathcal{G}=(\mathcal{X},\mathcal{E})$ consist of a set of atoms $\mathcal{X}=\{x_1, ..., x_i, ..., x_N\}$ and bonds $\mathcal{E}=\{e_{ij} | x_i\in \mathcal{X}, j\in N(i)\}$. $N(i)$ denotes the set of indices of the connected nodes of node $x_i$. To extract information from the molecular graph, message passing mechanism in a five-layered GIN graph encoder can be defined as:

\begin{equation}
z^{(k+1)}_i = \mathrm{MLP}^{(k+1)}_{\text{atom}} \left( z^{(k)}_i + \sum_{j \in N(i)} \left( z^{(k)}_j + \mathrm{MLP}^{(k+1)}_{\text{bond}}(e_{ij}) \right) \right)
\end{equation}

where \(k\in \{0, 1, 2, 3, 4\}\) and \( z^{0}_i = x_i \). 

The output of the five-layered GIN graph encoder follows linear transformation to convert the size of the dimension same as SMILES token embeddings. In addition, length truncation or padding follows to fix the length of the graph embeddings into $l$. Final graph representation $\mathcal{G}^*=(\mathcal{Z},\mathcal{E})$, with $\mathcal{Z}=\{z_1, ..., z_i, ..., z_l\}$, $\mathcal{E}=\{e_{ij} | z_i\in \mathcal{Z}, j\in N(i)\}$, is used for cross-token attention.

\begin{table*}[t]
    \centering
    \resizebox{\textwidth}{!}{
    \begin{tabular}{cccccccc}
        \toprule
        % Model #Trainable params BLEU-2 BLEU-4 ROUGE-1 ROUGE-2 ROUGE-L METEOR
        Input & Model & BLEU-2 & BLEU-4 & METEOR & ROUGE-1 & ROUGE-2 & ROUGE-L \\
        \midrule
        \multirow{3}*{\textbf{1D SMILES}}
        & MolT5, $Small$ [2022] & 14.8 & 8.5 & 18.5 & 26.5 & 13.5 & 23.6 \\
        ~ & MolT5, $Base$ [2022] & 30.1 & 20.9 & 35.6 & 40.3 & 25.1 & 33.8 \\
        ~ & MolT5, $Large$ [2022] & 30.2 & 22.2 & 36.6 & 41.5 & 25.9 & 34.8 \\
        \midrule
        % \multirow{3}*{\textbf{1D SMILES $+$ 2D Graph}}
        % & MoMu, $Base$ & 252M & 30.2 & 21.5 & 34.2 & 40.5 & 25.1 & 34.4 \\
        % ~ & MoMu, $Large$ & 782M & 31.1 & 22.8 & 36.2 & 41.8 & 25.7 & 36.7 \\
        \textbf{1D SMILES} & MolCA, $Galac_{125M}$ [2023] & 31.9 & 24.3 & 41.6 & 47.3 & 33.9 & 43.2  \\
        $+$ & MolCA, $Galac_{1.3B}$ [2023] & 38.7 & 30.3 & 45.6 & 50.2 & 35.9 & 44.5  \\
        \textbf{2D Graph} & GraphT5 [Ours, 2024] & \textbf{43.8} & \textbf{37.5} & \textbf{48.7} & \textbf{53.9} & \textbf{41.0} & \textbf{49.2} \\
        \bottomrule
    \end{tabular}}
    \caption{The performance (\%) of molecule captioning on PubChem324k dataset. The compared baselines include single-modal approaches with 1D SMILES input and multi-modal approaches utilizing both 1D SMILES and 2D graphs. Numbers in boldface indicate the best performance.}
    \label{table:PubChem_MolCap}
\end{table*}

\begin{table*}[t]
    \centering
    \resizebox{\textwidth}{!}{
    \begin{tabular}{cccccccc}
        \toprule
        % Model #Trainable params BLEU-2 BLEU-4 ROUGE-1 ROUGE-2 ROUGE-L METEOR
        Input & Model & BLEU-2 & BLEU-4 & METEOR & ROUGE-1 & ROUGE-2 & ROUGE-L \\
        \midrule
        \multirow{5}*{\textbf{1D SMILES}}
        & MolT5, $Small$ [2022] & 51.9 & 43.6 & 55.1 & 62.0 & 46.9 & 56.3 \\
        ~ & MolT5, $Base$ [2022] & 54.0 & 45.7 & 56.9 & 63.4 & 48.5 & 57.8 \\
        ~ & MolT5, $Large$ [2022] & 59.4 & 50.8 & 61.4 & 65.4 & 51.0 & 59.4 \\
        ~ & Text+Chem T5 [2023] & 58.0 & 49.0	& 60.4 & 64.7 & 49.8 & 58.6 \\
        ~ & Text+Chem T5, $augm$ [2023] & 62.5 & 54.2 & 64.8 & 68.2 & 54.3 & 62.2 \\
        % ~ & BioT5 & 252M & 63.5 & 55.6 & 65.6 & 69.2 & 55.9 & 63.3 \\
        \midrule
        % \multirow{5}*{\textbf{1D SMILES $+$ 2D Graph}}
         & MoMu, $Base$ [2022] & 54.9 & 46.2 & 57.6 & - & - & 57.5 \\
        \textbf{1D SMILES} & MoMu, $Large$ [2022] & 59.9 & 51.5 & 59.7 & - & - & 59.3 \\
        $+$ & MolCA, $Galac_{125M}$ [2023] & 61.2 & 52.6 & 63.6 & 67.4 & 52.1 & 60.6 \\
        \textbf{2D Graph} & MolCA, $Galac_{1.3B}$ [2023] & 62.0 & 53.1 & 65.1 & 68.1 & 53.7 & 61.8 \\
         & GraphT5 [Ours, 2024] & \textbf{63.8} & \textbf{56.6} & 64.1 & 67.7 & 53.7 & 61.7 \\
        \bottomrule
    \end{tabular}}
    \caption{The performance (\%) of molecule captioning on ChEBI-20 dataset. The compared baselines include single-modal approaches with 1D SMILES input and multi-modal approaches utilizing both 1D SMILES and 2D graphs. Numbers in boldface indicate the best performance.}
    \label{table:ChEBI_MolCap}
\end{table*}

\subsection{Cross-Token Attention}\label{method:cross}
% \begin{itemize}
%     \item Explain the strong point of multi-modal cross-token attention between Graph and the SMILES token
%     \item A mathematical explanation of how cross-token attention is calculated.
% \end{itemize}
In this work, we propose Cross-Token Attention to leverage the 1D SMILES and 2D graph representations of the molecule. Atom nodes of graph embeddings and SMILES tokens can interact via cross attention mechanism as follows:

\begin{equation}
Q = \mathcal{Z}\cdot W_Q, \quad K = \mathcal{S}^*\cdot W_K, \quad V = \mathcal{S}^*\cdot W_V 
\end{equation}
\begin{equation}
CrossTokenAttn(\mathcal{G}^*, \mathcal{S}^*) = \mathrm{softmax}\left(\frac{QK^T}{\sqrt{d_k}}\right)V 
\end{equation}
\begin{equation}
H = CrossTokenAttn(\mathcal{G}^*, \mathcal{S}^*)    
\end{equation}

where \( W_Q \), \( W_K \), and \( W_V \) are learnable weight matrices for queries, keys, and values, respectively. \( H\) represents the output of the attention, capturing the enriched information from SMILES tokens to the graph atom nodes. Subsequently, \( H \) is refined through a Layer Normalization with skip connection, then the MLP layer to produce the final output as follows:
\begin{equation}
H'= \mathrm{LayerNorm}(H + \mathcal{Z})
\end{equation}
\begin{equation}
O_G = \mathrm{MLP}(H')
\end{equation}

Finally, $O_G$ not only implies atom-level node representations within the molecular graph but also encompasses token-level interaction between different modalities of the molecules, which are graph and SMILES representations. The decoding process takes input from both $O_{Gpool}$, where average pooling has been applied to $O_G$, and the original format of $O_G$.

% \newpage
% add methods
% \newpage

\section{Experiments}
In this section, we conduct experiments to validate the effectiveness of the proposed GraphT5. We train our model for two molecular language modeling tasks, which are molecule captioning and IUPAC name prediction. 

\subsection{Experimental Setting}
% \begin{itemize}
%     \item Dataset, Hyperparameter settings, GPU ...
% \end{itemize}

\subsubsection{Datasets}
We use PubChem324k dataset \cite{liu2023molca} and ChEBI-20 dataset \cite{edwards2021text2mol} to train our model for the molecule captioning task. 
PubChem324k dataset provides not only molecule and description pairs but also graph representation of the molecule, which can be directly used for 1D SMILES and 2D graph multi-modal representation inputs. We use PubChem324k dataset \cite{liu2023molca} to train our model for molecule captioning and IUPAC name prediction tasks.
In the case of the ChEBI-20 dataset, we use \texttt{smiles2graph} operation provided from OGB \cite{hu2020open} to extract graph representation from SMILES representation and conduct molecule captioning task.
Table \ref{table:dataset} shows statistics of the PubChem324k and ChEBI-20 datasets. 

\subsubsection{Baselines}
For experiments with the PubChem324k dataset, we compare our GraphT5 with MolT5 \cite{edwards-etal-2022-translation} which is a 1D SMILES approach for molecular language modeling and MolCA \cite{liu2023molca} that utilizes 2D graphs as well as 1D SMILES but lacks cross-token attention.
In case of experiments with ChEBI-20 dataset, we compare our GraphT5 with following baselines: MolT5 \cite{edwards-etal-2022-translation}, Text+Chem T5 \cite{10.5555/3618408.3618651}, 
% BioT5 \cite{pei2023biot5} 
that utilize 1D SMILES representation, and MoMu \cite{su2022molecular}, MolCA \cite{liu2023molca} that additionally utilize 2D graph representation, but without interaction between 1D SMILES and 2D graph.
Table \ref{table:modelsize} demonstrates the sizes of the compared models, including our GraphT5. 

\subsubsection{Evaluation Metrics}
We evaluate our GraphT5 with molecule captioning and IUPAC name prediction tasks with metrics that are often used in natural language processing for tasks such as machine translation or summarization. Metrics that are used to evaluate generated results are as follows:
\begin{itemize}
    \item \textbf{BLEU-2, BLEU-4} \cite{papineni2002bleu}: BLEU (Bilingual Evaluation Understudy) measures the similarity between reference text and the generated output from a language model, based on precision. BLEU-2 and BLEU-4 consider 2-gram (bigram) and 4-gram respectively, to compute the overlap between the reference and generated texts. 
    \item \textbf{METEOR} \cite{banerjee2005meteor}: METEOR (Metric for Evaluation of Translation with Explicit ORdering) considers both unigram precision and recall between reference text and the generated output from a language model. In addition, METEOR considers stemming and synonymy of the words in texts to be compared. Therefore, it is regarded as a more precise evaluation metric compared to BLEU.
    \item \textbf{ROUGE-1, ROUGE-2, ROUGE-L} \cite{lin2004rouge}: ROUGE (Recall-Oriented Understudy for Gisting Evaluation) measures the similarity between reference text and the generated output from a language model, based on recall. ROUGE-1 and ROUGE-2 consider 1-gram (unigram) and 2-gram (bigram) respectively, to compute the overlap between the reference and generated texts. ROUGE-L computes the recall-overlap of the longest common subsequence (LCS) between the reference and generated texts.
\end{itemize}

\subsubsection{Implementation Details}
GraphT5 is implemented and trained based on PyTorch \cite{paszke2019pytorch} and Huggingface transformers \cite{wolf2020transformers} and optimized with AdamW \cite{loshchilov2017decoupled}. To initialize our text encoder and decoder, we use trained parameters from Text+Chem T5 \cite{10.5555/3618408.3618651}. We adopt pretrained weights from GraphMVP \cite{liu2021pre} to initialize our graph encoder. 
All experiments are trained on a single NVIDIA RTX A6000 GPU.
Detailed hyperparameter settings can be found in Appendix \ref{appenidx:hyper}.

% epoch, lr, number of layers, embedding dim, 
\begin{table*}[t]
    \centering
    \resizebox{\textwidth}{!}{
    \begin{tabular}{cccccccc}
        \toprule
        % Model #Trainable params BLEU-2 BLEU-4 ROUGE-1 ROUGE-2 ROUGE-L METEOR
        Input & Model & BLEU-2 & BLEU-4 & METEOR & ROUGE-1 & ROUGE-2 & ROUGE-L \\
        \midrule
        \multirow{3}*{\textbf{1D SMILES}}
        & MolT5, $Small$ [2022] & 48.6 & 35.2 & 42.5 & 40.0 & 16.1 & 34.3 \\
        ~ & MolT5, $Base$ [2022] & 52.7 & 41.5 & 53.2 & 50.7 & 26.0 & 44.3 \\
        ~ & MolT5, $Large$ [2022] & 59.4 & 49.7 & 58.5 & 55.9 & 33.3 & 49.1\\
        \midrule
        % \multirow{3}*{\textbf{1D SMILES $+$ 2D Graph}}
        \textbf{1D SMILES} & MolCA, $Galac_{125M}$ [2023] & 73.9 & 66.3 & 71.8 & 69.0 & 47.8 & 63.2     \\
        $+$ & MolCA, $Galac_{1.3B}$ [2023] & 75.0 & 66.6 & 72.1 & 69.6 & 48.2 & 63.4   \\
        \textbf{2D Graph} & GraphT5 & \textbf{81.9} [Ours, 2024] & \textbf{75.2} & \textbf{80.3} & \textbf{76.3} & \textbf{56.5} & \textbf{70.2} \\
        \bottomrule
    \end{tabular}}
    \caption{The performance (\%) of IUPAC name prediction on PubChem324k dataset. The compared baselines include single-modal approaches with 1D SMILES input and multi-modal approaches utilizing both 1D SMILES and 2D graphs. Numbers in boldface indicate the best performance. }
    \label{table:PubChem_IUPAC}
\end{table*}

\begin{table*}[t]
    \centering
    \resizebox{\textwidth}{!}{
    \begin{tabular}{cccccccccc}
        \toprule
        Model & Graph & SMILES & Cross-Token Attention & BLEU-2 & BLEU-4 & METEOR & ROUGE-1 & ROUGE-2 & ROUGE-L \\
        \midrule
        GraphT5 & {\color{cadmiumred}\xmark} & {\color{AoGreen}\cmark} & {\color{cadmiumred}\xmark} & 56.6 & 48.3 & 57.6 & 61.7 &	46.3 & 55.6 \\
        GraphT5 & {\color{AoGreen}\cmark} & {\color{cadmiumred}\xmark} & {\color{cadmiumred}\xmark} & 56.0	& 48.2	& 56.9	& 62.0	& 46.6	& 56.1 \\
        GraphT5 & {\color{AoGreen}\cmark} & {\color{cadmiumred}\xmark} & {\color{AoGreen}\cmark} & 62.2	& 54.8	& 62.8	& 66.5	& 52.0	& 60.5 \\
        GraphT5 & {\color{AoGreen}\cmark} & {\color{AoGreen}\cmark} & {\color{cadmiumred}\xmark} & 63.3	& 56.1	& 63.8	& 67.5	& 53.3	& 61.5 \\
        GraphT5 & {\color{AoGreen}\cmark} & {\color{AoGreen}\cmark} & {\color{AoGreen}\cmark} & \textbf{63.8} & \textbf{56.6} & \textbf{64.1} & \textbf{67.7} & \textbf{53.7} & \textbf{61.7} \\
        \bottomrule
    \end{tabular}}
    \caption{The performance results (\%) of the ablation study for different input modalities and cross-token attention, with molecule captioning task on ChEBI-20 dataset. `Graph' and `SMILES' indicate whether 2D graph or 1D SMILES embedding is included in the input for the decoder. `Cross-Token Attention' denotes if the proposed cross-token attention is applied to the graph embeddings for the decoder input. It can be verified that the multi-modal input and cross-token attention contribute to the superior performance of GraphT5.}
    \label{table:ablation}
\end{table*}

\subsection{Molecule Captioning}
% 2 Datasets why?
% PubChem more rich and high quality dataset : from various resources (including ChEBI-20), long descriptions
% Dataset statistics
% small dataset not enough to fully evaluate our model
% use 2 datasets to complement each other
\begin{figure}[t]
\centering
\includegraphics[width=0.95\columnwidth]{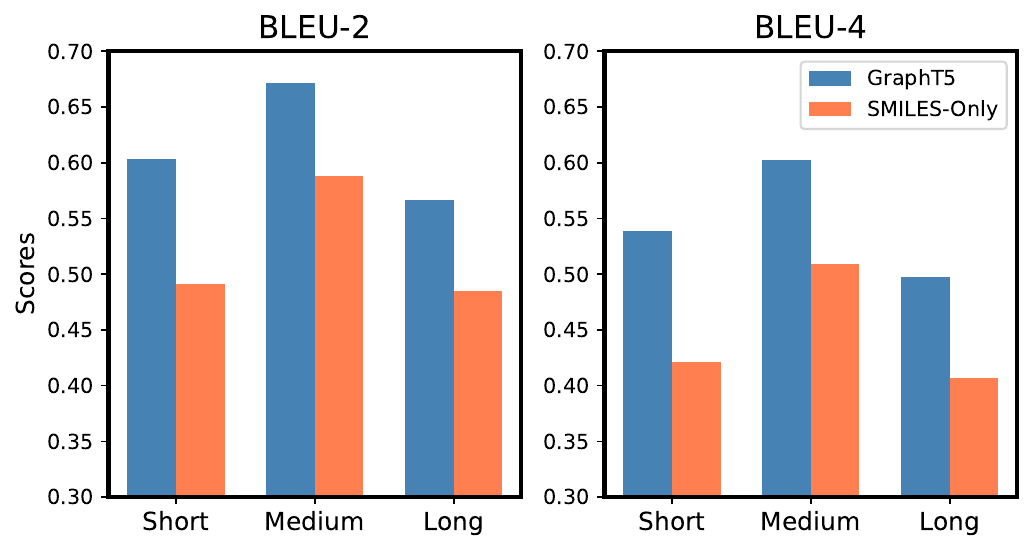} 
\caption{BLEU-2 and BLUE-4 score results for GraphT5 and 1D SMILES approach without graph utilization. The generated captions are evaluated in three groups, as divided by the length of the original description. Therefore, the robustness of the model to the lengths of the molecule descriptions can be validated.}
\label{fig:lengths}
\end{figure}

\subsubsection{Molecule Captioning Performance Result}
Molecule captioning task \cite{edwards-etal-2022-translation} aims to generate a textual description of a given molecule, as shown in Figure \ref{fig:summary}. The molecular input could be represented as 1D SMILES \cite{weininger1988smiles}, 2D graph, or both. We evaluate our GraphT5 for molecule captioning task, by training and evaluating with PubChem324k \cite{liu2023molca} and ChEBI-20 \cite{edwards2021text2mol} datasets respectively. 
By conducting experiments with two datasets which can verify the robustness along the datasets, we can alleviate concerns arising from the use of a single, small dataset.  
As shown in Table \ref{table:PubChem_MolCap}, GraphT5 shows superior performance compared to the baselines with PubChem324k dataset. 
Moreover, Table \ref{table:ChEBI_MolCap} shows that GraphT5 outperforms the baselines for BLEU-2 and BLEU-4 scores, and achieves comparable results for the other scores with the ChEBI-20 dataset.
As the PubChem324k dataset is constructed by collecting molecule descriptions from diverse sources, including ChEBI \cite{hastings2016chebi}, it is considered as containing longer and more accurate description of a given molecule compared to the ChEBI-20 dataset. Therefore, it is promising that our GraphT5 considerably outperforms the baselines, especially with the PubChem324k dataset.
According to the performance results, it has been verified that cross-token attention which extracts interactions between 1D SMILES and 2D graph representations could benefit the model without significantly increasing the size of the model.

\subsubsection{Robustness to Lengths}\label{result:robust}
We divide the ChEBI-20 test dataset into three groups by lengths, which are short, medium, and long (Appendix \ref{appendix:hist}). Each group is composed of 1,149, 1,076, and 1,075 pairs of molecules and their descriptions. Figure \ref{fig:lengths} shows performance results depending on the length of the molecule descriptions, where our GraphT5 outperforms the 1D SMILES approach regardless of the description length. The results demonstrate that our GraphT5 can effectively address molecular language modeling, even in the case that requires relatively longer caption generation.
% TODO : other metrics' bar graph to appendix

\subsection{IUPAC Name Prediction}
\begin{figure}[t]
\centering
\includegraphics[width=0.95\columnwidth]{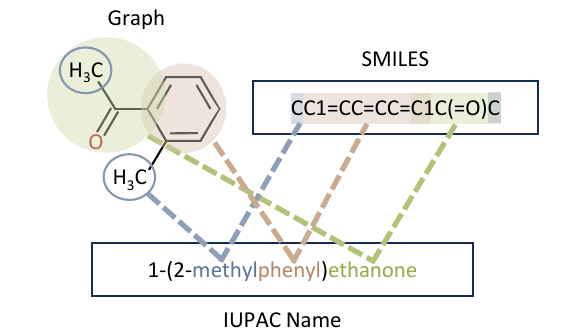} 
\caption{IUPAC name of a molecule reflects the structural characteristics of the molecule. The highlighted regions of the graph and SMILES representations stand for the same colored part of the IUPAC name.}
\label{fig:iupac}
\end{figure}
IUPAC name prediction task \cite{taylor2022galactica} aims to predict the IUPAC name of a given molecule. IUPAC stands for the International Union of Pure and Applied Chemistry, where a standardized naming system, IUPAC name, has been established \cite{iupac2005nomenclature}. Since the IUPAC naming system considers the structure of the molecule, for example, the longest carbon chain and the substituents, successful IUPAC name prediction requires sufficient understanding of the molecular structures. Figure \ref{fig:iupac} shows how the IUPAC name is related to the structure of a molecule, such as functional groups. Table \ref{table:PubChem_IUPAC} shows that our GraphT5 outperforms the baselines, with train and evaluation on the PubChem324k dataset.

% \subsection{Molecular Property Prediction}
\begin{figure*}[t]
\centering
\includegraphics[width=1\textwidth]{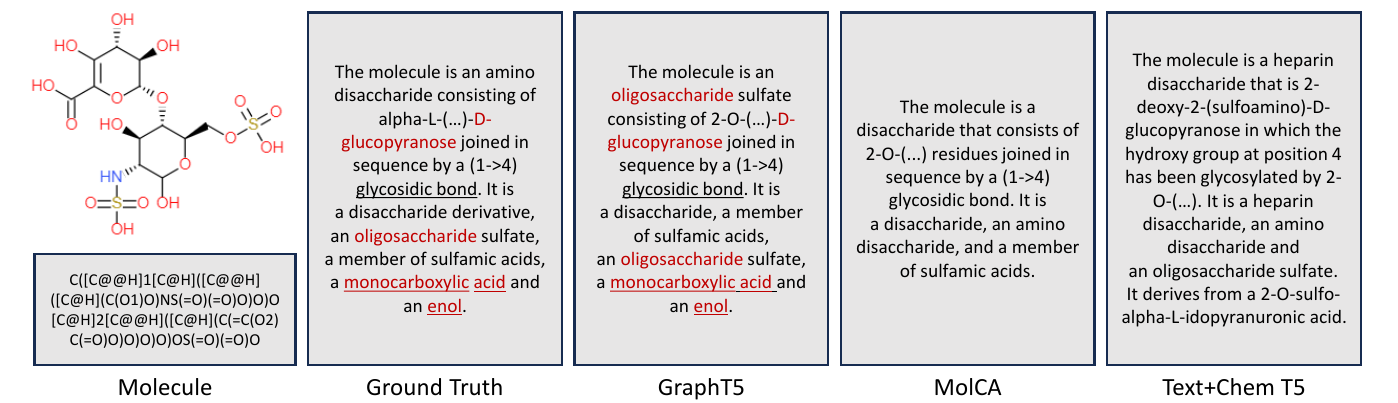} 
\caption{Examples of generated results from molecule captioning task. We compare our GraphT5 result with Text+Chem T5 which is a 1D SMILES text-based approach, and MolCA which utilizes 2D graphs as well as 1D SMILES but lacks cross-token attention. 
We highlight the correctly generated parts from our GraphT5 which cannot be found in MolCA with red color, and in the case of Text+Chem T5, those parts are underlined.
Since the generated captions are long, some of the middle parts that mostly share the contents among the examples are replaced by `(...)'.
}
\label{fig:casestudy}
\end{figure*}

\subsection{Ablation Studies}
% \begin{itemize}
%     \item Ablation study shows that using graph tokens and cross-token attention actually improves performance.
% \end{itemize}
To verify the effectiveness of multi-modal input (e.g. 1D SMILES and 2D graph) and cross-token attention proposed in GraphT5, we ablate the two conditions. The ablation studies are conducted with the ChEBI-20 dataset, and trained for molecule captioning tasks. Table \ref{table:ablation} demonstrates that appropriately combining 1D SMILES with 2D graphs for the input of the decoder is more beneficial than using only SMILES or graphs. 
Here, since the proposed cross-token attention is a method exclusively applied to graph embeddings, cross-token attention is not considered in the SMILES-only approach.
With the addition of SMILES information to graph representations through cross-token attention, the model's performance improves compared to relying solely on graph representations as shown in Table \ref{table:ablation}.
Therefore, it can be inferred that applying cross-token attention between 1D SMILES and 2D graph representations, which extracts interaction in fine-grained token level, successfully enriches the graph representation and leads to performance improvement in GraphT5.
Furthermore, as utilizing both 1D SMILES input and cross-token attention along with the graph representations yields the best performance results, cross-token attention is validated to further provide additional information that can be only extracted from the interaction of the two different modalities. 

% both approaches are validated to benefit the model.
% In summary, Table \ref{table:ablation} shows that the multi-modal input (e.g. 1D SMILES and 2D graph) structure is effective and the cross-token attention further provides additional information that can be only extracted from the interaction of the two different modalities. 

% On the other hand, the advantages of utilizing additional graphs with SMILES representations compared to using only SMILES can be inferred from Table \ref{table:ChEBI_MolCap}. 

\subsection{Case Study}
In this section, we examine how our GraphT5 generates the output caption for a given molecule, as shown in Figure \ref{fig:casestudy}. We conduct a comparative analysis with baseline models, such as Text+Chem T5, which utilizes only 1D SMILES, and MolCA, which combines 2D graphs with 1D SMILES but lacks cross-token attention. We use the ChEBI-20 dataset to train the models and generate examples. As depicted in Figure \ref{fig:casestudy}, the output of our GraphT5 contains more accurate and rich information compared to other baselines. 
For example, the generated output of our GraphT5 contains the term `monocarboxylic acid', which is present in the ground truth but absent in the outputs of MolCA and Text+Chem T5. Since `monocarboxylic acid' refers to a molecule structure with just one carboxylic acid group (COOH) \cite{YAO1998343}, it can be inferred that GraphT5 successfully captures the substructure of the given molecule. 

In the case of comparing the 2D graph and 1D SMILES approach but without cross-token attention (e.g. MolCA) with the 1D SMILES approach (e.g. Text+Chem T5), the result from using only 1D SMILES is composed of more extensive terms. However, terms are not always accurately generated in the 1D SMILES approach. For example, the term `heparin' generated from Text+Chem T5 refers to one of the anticoagulant medications with chemical formula $C_{12}H_{19}NO_{20}S_3$ \cite{hirsh1991heparin} which does not align with the given SMILES.
Therefore, it can be summarized that utilizing 2D graphs provides additional information about the molecule compared to using only 1D SMILES, and cross-token attention further extracts the interactions between 1D SMILES and 2D graph representations of the molecule that benefits successful molecular language modeling.
Words that are exclusively generated correctly by GraphT5 are highlighted in red or underlined, in contrast to MolCA and Text+Chem T5 respectively.

\section{Conclusion and Future Work}
In this work, we present GraphT5, a novel multi-modal molecular language modeling approach. GraphT5 utilizes both 1D SMILES and 2D graph representations of molecules to incorporate structural knowledge from these two modalities. Moreover, we introduce a novel cross-token attention method in GraphT5 to bridge the gap between the two different modalities of molecule representations, such as 1D SMILES and 2D graphs. 
Cross-token attention facilitates interaction and attention between 1D SMILES tokens and 2D graph nodes, which both have fine granularity, leading to enrichment in molecular structural knowledge. Extensive experiments on two datasets (e.g. PubChem324k,  ChEBI-20) and two tasks (e.g. Molecule captioning, IUPAC name prediction) validate the effectiveness of our GraphT5 with cross-token attention, outperforming several state-of-the-art baseline approaches.

For future work, we look forward to extending our proposed approach to aid drug discovery or material synthesis studies by providing detailed information about molecule structures and properties. Additionally, we expect further future work to explore the application of our approach in utilizing 3D molecular geometries, including molecular topology and 3D conformers.

\begin{acks}
To Robert, for the bagels and explaining CMYK and color spaces.
\end{acks}

\bibliographystyle{ACM-Reference-Format}
\bibliography{sample-base}

\appendix
% \section{Generated Molecule Captioning Samples}
% \begin{itemize}
%     \item smiles, gt, grapht5, molca, textchemt5
% \end{itemize}

\section{Histogram of number of words}\label{appendix:hist}

\begin{figure}[t]
\centering
\includegraphics[width=1\columnwidth]{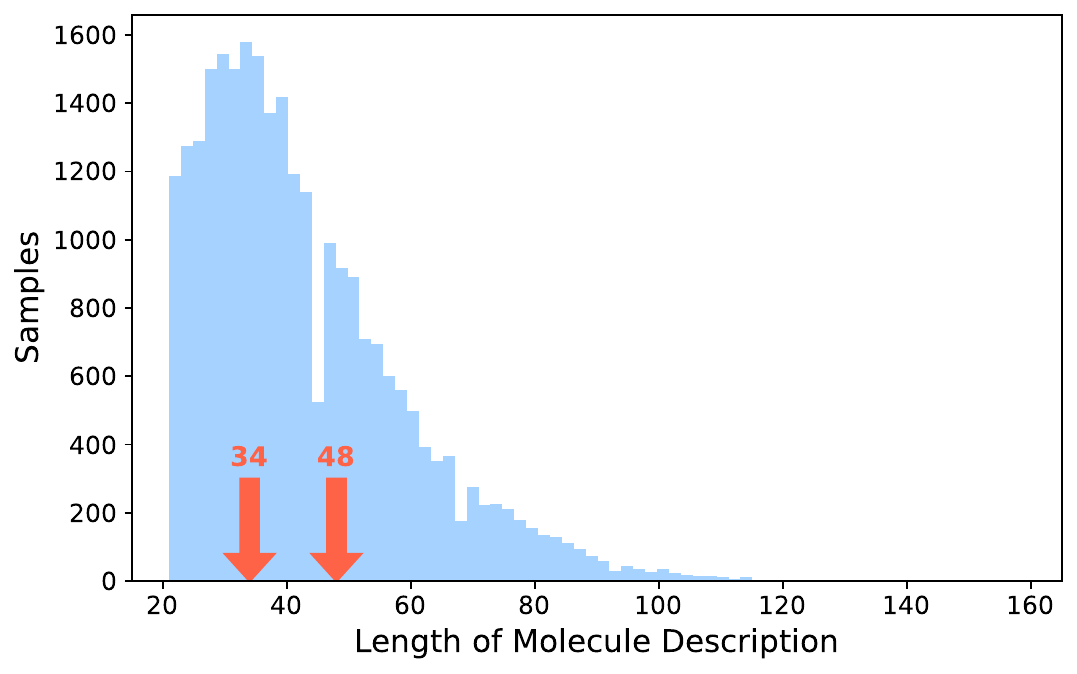} 
\caption{ Histogram for the length of molecule descriptions in train set. 34 and 48 are selected as
boundaries to split test data into three groups
(short, medium, and long), by the length of the description.}
\label{fig:hist}
\end{figure}

Figure \ref{fig:hist} shows the histogram for the number of words (tokens) per molecule description in ChEBI-20 \cite{edwards2021text2mol} train dataset. We select 34 and 48 as the boundaries to divide the test set into 3 groups by lengths (short, medium, and long), according to the first, second, and third quantile of the train data. We compare performance in three data subsets respectively, to validate our GraphT5's robustness to lengths (Section \ref{result:robust}).

\section{Prompt Templates}
Table \ref{table:prompt} shows prompt templates used for molecule captioning and IUPAC name prediction tasks.

\begin{table}[t]
    \centering
    \begin{tabular}{c|c}
        \toprule
        Task & Prompt Template \\
        \midrule
        Molecule Captioning & Caption the following molecule: \\
        ~ IUPAC Name Prediction & Predict IUPAC name of the following molecule: \\
        \bottomrule
    \end{tabular}
    \caption{Prompt templates used for molecule captioning and IUPAC name prediction tasks.}
    \label{table:prompt}
\end{table}

\section{Hyperparameter Settings}\label{appenidx:hyper}
Table \ref{table:hyperparams} shows hyperparameter search space during experiments. Underlined numbers indicate the optimized settings, which are used for final results.
\begin{table}[t]
    \centering
    % \resizebox{\textwidth}{!}{
    \begin{tabular}{cc}
        \toprule
        % Model #Trainable params BLEU-2 BLEU-4 ROUGE-1 ROUGE-2 ROUGE-L METEOR
        Hyperparameter & Search Space \\
        \midrule
        Epochs & 50, 80, 100, \underline{120}, 140, 150 \\
        Learning rate & 1e-3, \underline{1e-4}, 5e-5, 1e-5 \\
        Batch size & 10, \underline{14} \\
        Dropout ratio & 0.0, \underline{0.1}, 0.3 \\
        Attention Layers & \underline{12} \\
        Attention Heads & \underline{12} \\
        \bottomrule
    \end{tabular}
    \caption{Hyperparameter search space during experiments. Underlined numbers indicate the selected hyperparameters for the final results.}
    \label{table:hyperparams}
\end{table}
\end{document}